\documentclass[10pt,twocolumn,letterpaper]{article}

\usepackage{titling}

\usepackage{iccv}
\usepackage{times}
\usepackage{epsfig}
\usepackage{graphicx}
\usepackage{bm}
\usepackage{amsmath}
\usepackage{amssymb}
\usepackage{comment}
\usepackage{xcolor}
\usepackage{booktabs,makecell,multirow}
\usepackage{setspace}
\usepackage{threeparttable}
\usepackage{colortbl}
\usepackage{makecell}
\usepackage{wasysym}
\usepackage{authblk}
\usepackage[ruled,linesnumbered]{algorithm2e}
\usepackage[T1]{fontenc}
\definecolor{dark-gray}{gray}{0.40}
\definecolor{light-gray}{gray}{0.9}

\newtheorem{theorem}{Theorem}
\newtheorem{proposition}[theorem]{Proposition}

\usepackage[breaklinks=true,bookmarks=false]{hyperref}

\iccvfinalcopy 

\def\iccvPaperID{10507} 

\ificcvfinal\fi

\begin{document}

\makeatletter
\renewcommand\AB@affilsepx{\ \ \  \protect\Affilfont}
\makeatother

\title{\vspace{-7mm} Seeing Unseen: Discover Novel Biomedical Concepts via Geometry-\\Constrained Probabilistic Modeling
	\vspace{-5mm}}

\author[1]{Jianan Fan}
\author[1]{Dongnan Liu}
\author[2]{Hang Chang}
\author[3]{Heng Huang}
\author[4]{Mei Chen}
\author[1]{Weidong Cai\vspace{-2mm}}
\affil[1]{\normalsize University of Sydney} 
\affil[2]{\normalsize Lawrence Berkeley National Laboratory}
\affil[3]{\normalsize University of Maryland at College Park}
\affil[4]{\normalsize Microsoft}
\affil[ ]{\small \texttt{jfan6480@uni.sydney.edu.au\qquad dongnan.liu@sydney.edu.au\qquad  hchang@lbl.gov\qquad henghuanghh@gmail.com\qquad Mei.Chen@microsoft.com\qquad tom.cai@sydney.edu.au}}

\maketitle
\ificcvfinal\thispagestyle{empty}\fi

\begin{abstract}
	Machine learning holds tremendous promise for transforming the fundamental practice of scientific discovery by virtue of its data-driven nature.
	With the ever-increasing stream of research data collection, it would be appealing to autonomously explore patterns and insights from observational data for discovering novel classes of phenotypes and concepts.
	However, in the biomedical domain, there are several challenges inherently presented in the cumulated data which hamper the progress of novel class discovery.
	The non-i.i.d. data distribution accompanied by the severe imbalance among different groups of classes essentially leads to ambiguous and biased semantic representations.
	In this work, we present a geometry-constrained probabilistic modeling treatment to resolve the identified issues.
	First, we propose to parameterize the approximated posterior of instance embedding as a marginal von Mises-Fisher distribution to account for the interference of distributional latent bias.
	Then, we incorporate a suite of critical geometric properties to impose proper constraints on the layout of constructed embedding space, which in turn minimizes the uncontrollable risk for unknown class learning and structuring.
	Furthermore, a spectral graph-theoretic method is devised to estimate the number of potential novel classes.
	It inherits two intriguing merits compared to existent approaches, namely high computational efficiency and flexibility for taxonomy-adaptive estimation.
	Extensive experiments across various biomedical scenarios substantiate the effectiveness and general applicability of our method.
	\vspace{-1.5mm}
\end{abstract}

\section{Introduction}
\label{sec:introduction}
The rapid progress in deep learning\;(DL) has invigorated enormous interests for automating essential stages in biomedical image analysis through data-driven methodologies \cite{chen2020computer, song2021visual}.
However, the transformative potential of DL to facilitate scientific discovery at the fundamental level remains largely untapped.
\begin{figure}[!t]
	\centerline{\includegraphics[width=\columnwidth]{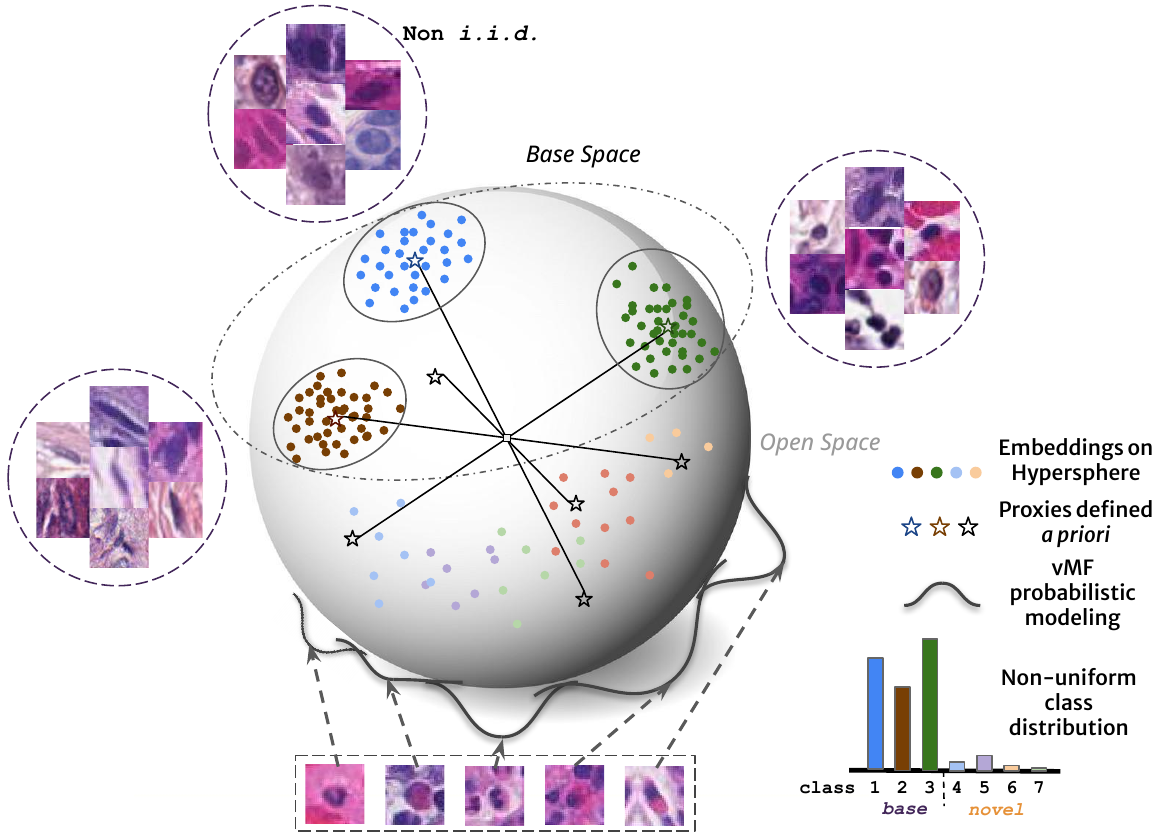}}
	\caption{Conceptual illustration of the main insight. In the biomedical domain, violation of the i.i.d. assumption incurred by inconsistent imaging protocols across cohorts and non-uniform class distributions due to scarcity of rare classes could deteriorate the generalizability of learned representations for novel class discovery.
		We propose to address those issues via probabilistic modeling on a hyperspherical manifold and incorporation of geometrical inductive biases for countering semantic ambiguity and open space risk.}
	\label{fig:insight}
	\vspace{-3.7mm}
\end{figure}
Specifically, the success of existing works hinges on expert-curated data annotations in which associated biomedical structures\;(e.g., cells, tissues) and disease categories are defined a priori \cite{liu2021panoptic, fan2024learning}.
They have little to offer for 
exploring unknown concepts and discovering expressive profiles for novel/undefined biomedical phenotypes from observational data alone.
For instance, cell, the basic unit of human biological system, exhibits vast diversity with hundreds of different types and states \cite{macosko2015highly}.
Compared with cumbersome manual pattern identification and grouping of cell populations, autonomously mining new classes of cells with their morphological and functional characteristics could greatly speed up the scientific discovery progress of biomedical research.

To this end, we study the problem of discovering novel concepts from biomedical imaging data, under the \emph{Generalized Novel Class Discovery} setting \cite{vaze2022generalized}.
Given a reference dataset within a set of well-defined base categories, it aims at identifying and clustering unseen classes in an external collection of image instances without supervision \cite{han2019learning, Han2020Automatically}.
The technical challenge of this task lies in how to leverage the structured prior knowledge secured from previously known biomedical object categories to guide the unsupervised discovery of new concepts.
Recently, a stream of methods based on representation learning and clustering \cite{fini2021unified, han2021autonovel, vaze2022generalized} have been proposed to detect 
unseen objects in generic visual perception scenes.
However, the basic assumption those works built upon may not hold in the biomedical domain from two aspects:
\textbf{i)} It is generally assumed that the labeled and unlabeled data are independent and identically distributed\;(i.i.d.) \cite{zheng2022towards, joseph2022novel}, whereas for biomedical images, data distribution shifts inevitably arise from the variations in targeted biology and image acquisition among multiple cohorts \cite{liu2021feddg, fan2023taxonomy}.
The statistical discrepancy between the reference set of data and the external collection for discovery would consequently result in ambiguous semantic representations given the interference of task-irrelevant attributes \cite{liu2020unsupervised}.
\textbf{ii)} Existing works hypothesize a balanced class distribution a priori which expects that the quantity and occurrence frequency of unseen classes should be on par with the base classes \cite{chi2022meta, zhao2022novel}.
They dismiss the long-tail nature of class distribution in scientific discovery that instances of classes never seen before are essentially rare and hard to collect.
For instance, rare diseases are inherently atypical and much fewer in caseload than their common counterpart \cite{codella2019skin, marrakchi2021fighting}.
The formulated embedding space is consequently prone to be biased towards the base classes with dominatedly copious training samples.
Despite the latest efforts delivering explicit modeling of the distribution biases \cite{yu2022self, yang2023bootstrap}, their self-training founded solutions intrinsically fall short due to erratic representation space structuring.
Without consideration of the dark information \cite{chen2020learning} regarding unknown class representations, there is no guarantee that a structured open space can be preserved for learning new classes.
It deteriorates the generalizability of those methods to uncover and group the implicit patterns of unseen phenotypes.

In this work, we present a geometry-constrained probabilistic treatment for discovering novel biomedical classes which resolves the above-mentioned challenges from a new perspective, as illustrated in Fig.\;\ref{fig:insight}.
First, to counteract the semantic ambiguities induced by data distribution shifts, we suggest to circumvent deterministic image encoding and instead model latent representations as directional distributions.
Specifically, for each instance, we parameterize the approximated posterior of its semantic embedding with a marginal von Mises-Fisher\;(vMF) distribution on a unit hyperspherical manifold.
The probabilistic modeling presents an elegant solution to decouple latent bias incurred by inconsistent imaging protocols from informative semantic context as an implicit uncertainty measure.
Second, we pinpoint the necessity to incorporate \emph{inductive biases} for imposing proper constraints on the geometric layout of learned embedding space and thereby minimizing the uncontrollable risk for unknown classes.
Aside from the \emph{hyperspherical manifold} prior, we identify two critical geometric properties for embedding space structuring, i.e., \emph{boundness} and \emph{uniformity}.
Here, \emph{boundness} implies that, in the latent space, both base and novel classes should be limited to their certain bounded ranges with a large margin separation between the two groups.
\emph{Uniformity} refers to the degree of distributional diversity for the set of semantic proxies on the hypersphere.
These principles cooperatively shape the layout of embedding space towards risk-bounded novel class discovery and perspicuous geometric interpretation.
With explicit constraints to integrate the underlined inductive biases into representation learning, our proposed method inherits guarantees to attain maximal inter-class separability and intra-class compactness, which in turn addresses the inordinate dominance of base classes and favors the generalization of learned embeddings to recognize unseen concepts.

Additionally, with respect to the practice of existing works assuming the number of novel classes are known a priori \cite{zhong2021neighborhood, li2023modeling}, we argue that such supposition is over-optimistic and unrealistic in the open world of scientific discovery.
We therefore propose a versatile and plug-and-play solution to estimate the number of categories in unlabeled data.
The spectral graph theory-based approach is computationally efficient and does not introduce recursive clustering as in \cite{vaze2022generalized}.
It also holds an appealing merit that it supports taxonomy-adaptive estimation, which means the users can decide the granularity level\;(e.g., coarse or fine-grained) in class hierarchy with regards to their specific needs and get the corresponding outcome.

\noindent\textbf{Contributions.} \ 
Our contributions are four-fold:
\textbf{(i)}\;We introduce a novel paradigm to unleash the tremendous promise of DL to facilitate data-driven biomedical discovery and account for the distribution biases inherently presented in the domain.\;\textbf{(ii)}\;We propose geometry-constrained probabilistic modeling to facilitate risk-bounded novel class discovery by exploiting the \emph{boundness} and \emph{uniformity} geometrical inductive biases for representation space structuring.\;We perform theoretical justifications from two perspectives to illustrate the merits of our method.
\textbf{(iii)}\;We devise a spectral graph-theoretic method to estimate the potential number of novel classes in unlabeled observational data.
\textbf{(iv)}\;
The proposed method is extensively validated on a diverse suite of challenging scenarios and attains superior performance compared to existing solutions.

\section{Related Works}
\noindent\textbf{Novel Class Discovery.} 
Inspired by the observation that human visual system can effortlessly recognize an unseen class of objects based on previously learnt category concepts, a stream of methods have been proposed for novel class discovery\;(NCD) in an unsupervised manner by transferring prior knowledge gathered from a set of seen categories with annotations \cite{han2019learning, Han2020Automatically, han2021autonovel}.
Follow-up works further identify the limitations of NCD in that it disregards the compositional nature of real-world upcoming data and supposes encountered classes will not appear again in the unlabeled set.
To this end, they formalize the \emph{Generalized Novel Class Discovery}\;(GNCD) setting which assumes both base and novel classes could simultaneously appear in the unlabeled data \cite{vaze2022generalized}.
ComEx \cite{yang2022divide} and DPN \cite{an2023generalized} propose to decouple the two sets of categories for capturing discriminative representations.
In a similar spirit, PromptCAL \cite{zhang2023promptcal} and DCCL \cite{pu2023dynamic} resort to synergistic learning and seek to improve the constructed representations via auxiliary training objectives.
Nevertheless, those works are rooted in over-optimistic assumptions of i.i.d. data and balanced class distribution \cite{vaze2022generalized} and thereby tend to suffer from severe learning bias in real-world scientific discovery with intricate circumstances.
Despite the recent efforts to account for non-i.i.d.\;and non-uniform data distribution in consonance with NCD \cite{yu2022self, yang2023bootstrap}, their solutions are limited to the commonly adopted self-training scheme.
Without explicit regularizations imposed on the geometric layout structure of the embedding space, their performance could fluctuate vastly given the volatile quality of pseudo-labels \cite{wang2020unsupervised}.

\noindent\textbf{Open-World Semi-Supervised Learning.} \ 
The technical challenge we intend to unravel is in line with an arising topic, i.e., open-world semi-supervised learning \cite{cao2022openworld}.
It takes a step further than the closed-world setting held in standard semi-supervised learning and presumes the emergence of new classes in unlabeled training and test dataset.
Following the insight, several recent works suggest the usage of pseudo-labels and pair-wise similarity to group and recognize unseen categories \cite{rizve2022openldn, cao2022pss, rizve2022towards, guo2022robust}.
However, their promise in biomedical scientific discovery is unfavorably undermined by the challenges posed by distribution bias.


\noindent\textbf{von Mises-Fisher Distribution.} \ 
Probabilistic representation learning provides an elegant viewpoint to measure predictive uncertainty and mitigate feature ambiguity arising from deterministic mappings \cite{shi2019probabilistic}.
The multivariate Gaussian distribution is commonly employed as a prominent choice for probabilistic modeling in Euclidean space, whilst its applicability on spherical embeddings is limited due to inherent conflicts with the underlying manifold geometry \cite{davidson2018hypersphe}.
On the contrary, the von Mises-Fisher\;(vMF) distribution is introduced as a directional statistical model that is defined on the hypersphere.
It has been applied in tasks including facial recognition \cite{li2021spherical, xu2023probabilistic}, out-of-distribution detection \cite{ming2023exploit}, and long-tailed learning \cite{wang2022towards}, leading to remarkable advancements in these areas.
Nevertheless, the significance of incorporating crucial inductive biases pertaining to geometric properties within the embedding space, such as uniformity, has been overlooked in the literature.
Distinguished from existing works, to the best of our knowledge, our study is the first to explicitly structure the geometric layout of learned representations and account for the statistical discrepancy across data with distributional concentration modeling.

\section{Methodology}
\subsection{Problem Statement}
The objective of this study is to achieve automated discovery of clusters that correspond to novel concepts using observational data, which is accomplished by transferring the knowledge acquired from a set of base classes with labeled image samples.
Following the practical setting in \cite{vaze2022generalized}, we assume that there exists a labeled base set $\mathcal{D_B}=\{(\bm{x}_i, \bm{y}_i)\}_{i=1}^{N_\mathcal{B}}\in\mathcal{X_B}\times\mathcal{Y_B}$,
and an unlabeled set $\mathcal{D_U}=\{(\bm{x}_i, \bm{y}_i)\}_{i=1}^{N_\mathcal{U}}\in\mathcal{X_U}\times\mathcal{Y_U}$ with potential novel classes, where $\mathcal{Y_U}$ are not available.
All instances in $\mathcal{D_B}$ are of class set $\mathcal{S_B}$.
As base classes also appear in the unlabeled set, i.e., $\mathcal{Y_B}\subset\mathcal{Y_U}$, the set of novel classes to be discovered can be formulated as $\mathcal{S}_N=\mathcal{Y_U}\backslash\mathcal{Y_B}$, with the total number $|\mathcal{S}_N|$ unknown a priori.
Notably, distinguished from \cite{vaze2022generalized}, the two sets of data are drawn from different distributions due to inconsistent biology processing and imaging protocols, i.e., $\mathcal{X_B}\sim\mathcal{P_B}, \mathcal{X_U}\sim\mathcal{P_U}, \mathcal{P_B}\neq\mathcal{P_U}$.
The class distribution of $\mathcal{D_U}$ is innately non-uniform as well.
Given a new instance from $\mathcal{P_U}$, the goal is to either recognize it as one of the base classes or find it is closest to which identified novel class amongst $\mathcal{S}_N$.

\subsection{Geometry-Constrained Probabilistic Modeling}\label{sec:mainmethod}
\begin{figure}[!t]
	\centerline{\includegraphics[width=1.1\columnwidth]{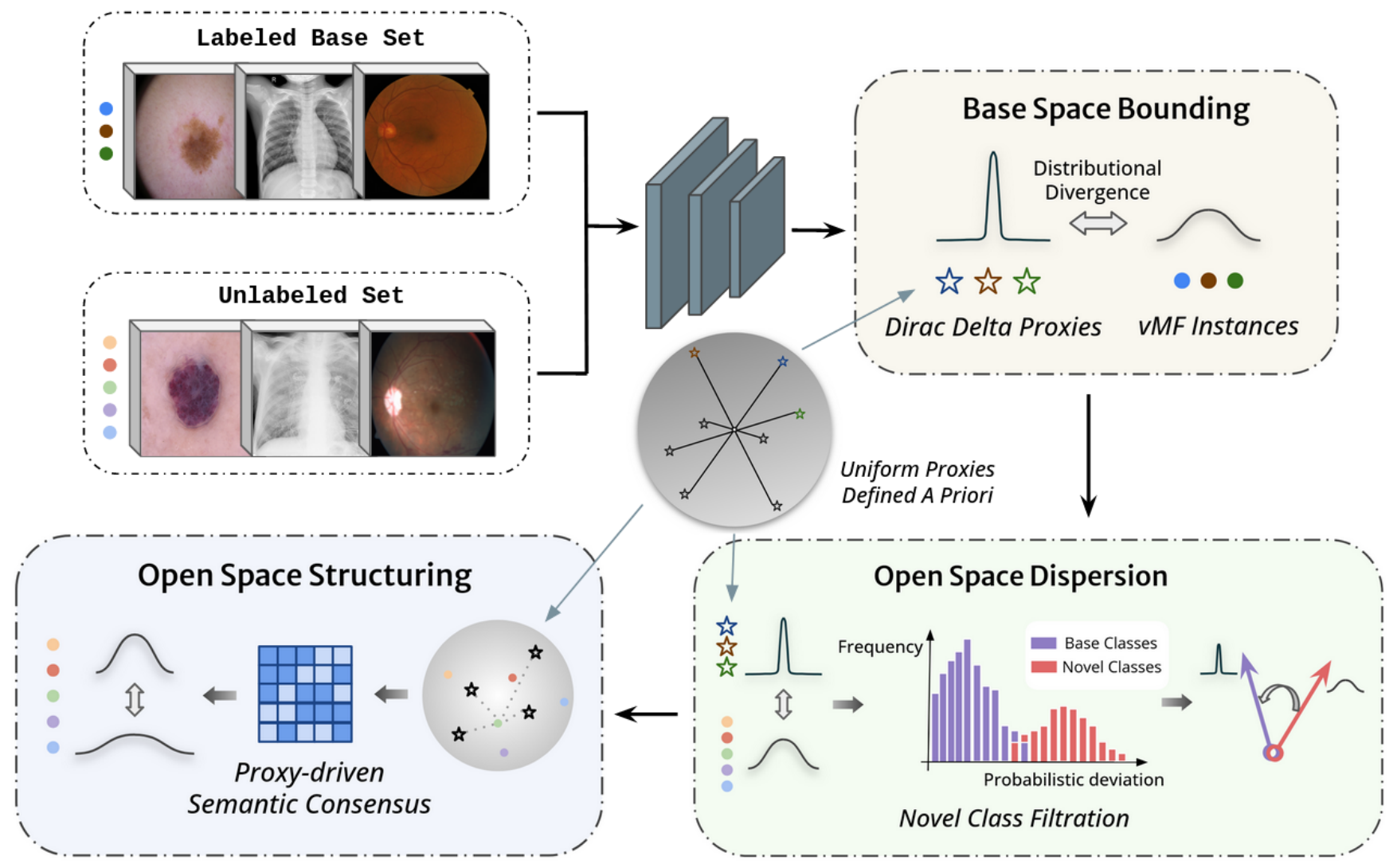}}
	\caption{Overview of the proposed method. We propose to incorporate the \emph{uniformity} and \emph{boundness} geometrical inductive biases by establishing preorganized proxies as anchors and then structuring the geometric layout of learned embedding space successively with hyperspherical probabilistic modeling.}
	\label{fig:paradigm}
\end{figure}

\noindent\textbf{Overview.} \ 
The technical challenge to adapt base knowledge for novel biomedical concept discovery lies in two aspects: semantic ambiguity caused by data distribution shifts and uncontrollable open space risk as a result of major class dominance.
To this end, we propose a novel geometry-constrained probabilistic treatment.
The framework is illustrated in Fig.\;\ref{fig:paradigm}.
It explicitly decouples the underlying distributional bias from informative semantic representations and structures the geometric layout of the formulated embedding space based upon inductive biases on latent manifold.
First, we incorporate the \emph{uniformity} inductive bias by positioning semantic proxies a priori with uniform distribution guarantees.
It helps to mitigate the representation collapse incurred by the inherent imbalance between base and novel classes.
Subsequently, we leverage vMF probabilistic modeling to promote latent space \emph{boundness} by firstly establishing a compact manifold for each base class and then regularizing the open space to be separated from the base manifold by a large angular margin. 
At last, to demystify the dark information regarding unknown class representations, we propose to structure the embedding space and minimize the open space risk by imposing statistical discrepancy constraints on unlabeled instances with semantic consensus.

\noindent\textbf{Uniform Proxies Founded A Priori with Energy Minimization.} \ 
Prior to representation learning, we propose to at first organize the embedding space with pre-defined proxies.
By disentangling proxy foundation as an individual step, we seamlessly integrate the \emph{uniformity} inductive bias to shape model output space without disturbing discriminative class learning.
Here, uniformity indicates the degree of distributional diversity for the set of proxies.
With the proxies distributed uniformly throughout the entire manifold, we minimize the open space risk by explicitly structuring the feature space for unknown classes.
The equidistributional organization also facilitates inter-class separation and intra-class compactness.
Specifically, given a \emph{d}-dimensional embedding space, we define a set of prior proxies $\bm{\hat{\Upsilon}}=\{\bm{\hat{\upsilon}}_1,...,\bm{\hat{\upsilon}}_n\}\in\mathbb{S}^{d-1}$ on a unit hyperspherical manifold.
The number of proxies $n$ should be generally larger than the total number of classes, for which sensitivity analysis is conducted in the supplementary.
We then propose to characterize the distributional uniformity of those proxies with their hyperspherical potential energy \cite{saff1997distributing}.
Energy behaves as a measure of distribution redundancy where lower energy implies the proxies are evenly positioned with large margin separation.
We thereby suggest to minimize the potential energy with the following objective:
\begin{equation}\label{equ:energy}
	\mathop{\mathtt{arg\;min}}\limits_{\{\bm{\hat{\upsilon}}_1,...,\bm{\hat{\upsilon}}_n\}\in\mathbb{S}^{d-1}}\,\bm{E}_{s,d}(\bm{\hat{\upsilon}}_i|_{i=1}^n)=\sum_{i=1}^{n}\sum_{j=1,j\neq i}^{n}\xi_{s,d}(\bm{\hat{\upsilon}}_i,\,\bm{\hat{\upsilon}}_j),
\end{equation}
where the unit hypersphere $\mathbb{S}^{d-1}=\{\bm{\hat{\upsilon}}\in\mathbb{R}^d |\,\lVert\bm{\hat{\upsilon}}\rVert=1\}$.
$\xi_{s,d}(\cdot,\,\cdot)$ represents the measurement of the correlation between proxies, which is typically implemented with the following Riesz \emph{s}-kernel function:
\begin{equation}\label{equ:kernel}
	\xi_{s,d}(\bm{\hat{\upsilon}}_i,\,\bm{\hat{\upsilon}}_j):= \left\{\begin{aligned}
		\Gamma(\bm{\hat{\upsilon}}_i,\,\bm{\hat{\upsilon}}_j)^{-s} &,\ \ s>0, \\
		\log(\Gamma(\bm{\hat{\upsilon}}_i,\,\bm{\hat{\upsilon}}_j)&^{-1}),\ \ s=0,
	\end{aligned}\right.
\end{equation}
where $\Gamma(\cdot,\,\cdot)$ is defined as the hyperspherical geodesic distance.

\noindent\textbf{Distributional Characterization on Hypersphere.} \ 
To account for the semantic ambiguities induced by data distribution shifts, we propose probabilistic modeling to parameterize latent embeddings as directional distributions on a hyperspherical manifold.
Specifically, we first let $f_\theta$ represent a projection network that maps input images to their hyperspherical embedding, i.e., $f_\theta:\bm{x}\in\mathcal{X}\mapsto\bm{z}\in\mathbb{S}^{d-1}$. 
Then, we characterize the approximated posterior of each instance embedding as a marginal von Mises-Fisher (vMF) distribution, i.e., $q_\theta(\bm{z}|\bm{x})\sim\mathbf{vMF}(\bm{\tilde{\mu}_x,\tilde{\kappa}_x})$.
Formally, a vMF distribution is defined as:
\begin{equation}\label{equ:vmf1}
	q(\bm{z}|\bm{\tilde{\mu}_x,\tilde{\kappa}_x})=C_d(\bm{\tilde{\kappa}_x})\exp(\bm{\tilde{\kappa}_x\tilde{\mu}_x}^T\bm{z}),
\end{equation}
\begin{equation}\label{equ:vmf2}
	C_d(\bm{\tilde{\kappa}_x})=\frac{\bm{\tilde{\kappa}_x}^{d/2-1}}{(2\pi)^{d/2}\mathcal{I}_{d/2-1}(\bm{\tilde{\kappa}_x})},
\end{equation}
where $\mathcal{I}_{d/2-1}(\cdot)$ denotes the modified Bessel function of the first kind at order $(d/2-1)$.
$\bm{\tilde{\mu}_x}$ and $\bm{\tilde{\kappa}_x}$ are statistical parameters representing the direction and concentration of the distribution.
With this formulation, the discriminative semantic context can be derived from the directional placement on the hypersphere and hence get decoupled from task-irrelevant variational attributes.
Besides, the concentration parameter $\bm{\tilde{\kappa}_x}$ provides an intuitive measurement of semantic embedding ambiguity \cite{scott2021mises} and can therefore act as an instance-wise temperature factor.
It adaptively scales the contributions of different image samples based on their representation uncertainty to deliver dynamic rectifications on the latent space structuring procedure, which is theoretically analyzed afterwards.

\noindent\textbf{Base Space Bounding and Open Space Dispersion.} \ 
With the modeled probabilistic representations, the subsequent objective is to organically structure the geometric layout of embedding space for discriminative novel class discovery.
Ideally, the space covered by well-defined base categories should be bounded within a certain range while the unknown classes corresponding to the open space should be modeled apart from the periphery of base space and dispersed over the entire manifold \cite{chen2020learning}.
We therefore propose to leverage the pre-defined proxies to bound the base space and keep the open space isolated.
For each base class, we select one proxy as the class anchor and reformulate it with Dirac delta $\delta$ to derive a statistical modeling over its deterministic representation:
\begin{equation}\label{equ:delta}
	\Delta(\bm{z}):=\left\{\begin{aligned}
		\int_{\mathbb{S}^{d-1}}\delta(\bm{z}-\bm{\hat{\upsilon}}_\mathcal{B})d\bm{z}&=1;\\
		\int_{\mathbb{S}^{d-1}}\delta(\bm{z}-\bm{\hat{\upsilon}}_\mathcal{B})\phi(\bm{z})d\bm{z}&=\phi(\bm{\hat{\upsilon}}_\mathcal{B}).
	\end{aligned}\right.\end{equation}
Here $\bm{\hat{\upsilon}}_\mathcal{B}$ denotes the fixed positional values of the proxies for base classes.
It is noted that the selection of proxies is random since they have been forced to be uniformly distributed on a symmetric manifold and all proxy pairs have the same distance.
Then, the proxies can behave as desired latent priors to regularize the distribution of base classes.
Specifically, the optimization objective is to enforce the instance-wise vMF distribution $q(\bm{z}|\bm{\tilde{\mu}_x,\tilde{\kappa}_x})$ approximating the latent prior $\Delta(\bm{z})$ in terms of KL divergence:
\begin{equation}\label{equ:kl_ilu}
	\mathop{\mathtt{min}}\limits_{q}\mathbb{E}_{\mathcal{D_B}}[-(\int_{\mathbb{S}^{d-1}}\Delta(\bm{z})\log q(\bm{z}|\bm{\tilde{\mu}_x,\tilde{\kappa}_x})d\bm{z})-\mathcal{H}_\Delta(\bm{z})],
\end{equation}
where $\mathcal{H}_\Delta(\bm{z})$ denotes the differential entropy over $\Delta(\bm{z})$, which is a constant as $\Delta(\bm{z})$ is fixed a priori.
Combined with Eq.\,(\ref{equ:vmf1})-(\ref{equ:delta}), the objective is equivalent to minimizing the following loss function:
\begin{equation}\label{equ:kl_loss}\begin{aligned}
		&\mathcal{L_B}(\bm{\tilde{\mu}_x,\tilde{\kappa}_x})=-\int_{\mathbb{S}^{d-1}}[\bm{\tilde{\kappa}_x\tilde{\mu}_x}^T\bm{z}+\log C_d(\bm{\tilde{\kappa}_x})]\Delta(\bm{z})d\bm{z} \\
		&=-\bm{\tilde{\kappa}_x\tilde{\mu}_x}^T\bm{\hat{\upsilon}}_\mathcal{B}-(d/2-1)\log\bm{\tilde{\kappa}_x}+\log\mathcal{I}_{d/2-1}(\bm{\tilde{\kappa}_x})+\check{\gamma},
\end{aligned}\end{equation}
where $\check{\gamma}=(d/2)\log2\pi$ is a constant constituent.
The regularization term constrains the intra-class variability and shrinks the base space within a bounded radius.
Thereafter, we propose open space dispersion aimed at separating the space reserved for novel classes away from the base one.
The challenging aspect is how to recognize the unseen concepts from the hybrid unlabeled set of data where base and novel classes concurrently exist.
In this regard, we first derive the magnitude of probabilistic deviation as an indication of open-set instances and hereby enforce large angular margins between those embeddings and the base space.
Specifically, we measure the probability density of each instance-wise vMF distribution with respect to base class proxies and estimate the ranking over $\mathcal{X_U}$:
\begin{equation}\label{equ:openset}
	\mathop{argsort}\limits_{\{\bm{x}_1,...,\bm{x}_{N_\mathcal{U}}\}\in\mathcal{X_U}}
	\Big\{\mathop{\mathtt{max}}\limits_{\{\bm{\hat{\upsilon}}_{\mathcal{B}_i},\forall i\in\mathcal{S_B}\}} \,
	C_d(\bm{\tilde{\kappa}_x})\exp(\bm{\tilde{\kappa}_x\tilde{\mu}_x}^T\bm{\hat{\upsilon}}_{\mathcal{B}_i})\Big\}.
\end{equation}
The batch-wise instances with maximal divergence from the base class proxies are hereafter pushed towards the open space:
\begin{equation}\label{equ:disperse}
	\mathcal{L}_{dis}(\bm{\tilde{\mu}_x,\tilde{\kappa}_x})=-\log\frac{\exp(\bm{\tilde{\kappa}_x\tilde{\mu}_x}^T\bm{\hat{\upsilon}}_\mathcal{R})}{\exp(\bm{\tilde{\kappa}_x\tilde{\mu}_x}^T\bm{\hat{\upsilon}}_\mathcal{R})+\exp(\bm{\tilde{\kappa}_x\tilde{\mu}_x}^T\bm{\hat{\upsilon}}_\mathcal{B})},
\end{equation}
where $\bm{\hat{\upsilon}}_\mathcal{R}=\bm{\hat{\Upsilon}}\setminus\bm{\hat{\upsilon}}_\mathcal{B}$ denotes the remaining proxies located outside of the base space.
The optimization objectives of Eq.\;(\ref{equ:kl_loss}) and (\ref{equ:disperse}) cooperatively promote \emph{boundness} on the underlying manifold.
We then propose to explicitly introduce the $uniformity$ inductive bias and accordingly structure the open space to control its associated risk, which in turn appreciates the generalizability of learned representations for recognizing unseen concepts.

\noindent\textbf{Open Space Structuring.} \ 
The dark information in the unknown space which has not been properly characterized raises uncontrollable risks for novel classes learning and discrimination \cite{chen2020learning}.
To this end, we propose to structure the open space organically with proxy-driven semantic consensus, which encourages the constructed embeddings to be uniformly distributed over the manifold with semantically-meaningful clusters such that the representations of different novel classes can be spaced apart with large angular distance.
Our intuition is that instances with adjacent geometric placements on the underlying manifold should possess analogous semantics and exhibit marginal distributional discrepancy.
Specifically, given each pair of unlabeled instances $(\bm{x}_\alpha,\bm{x}_\beta)$ and corresponding probabilistic modeling $q_\theta(\bm{z}_\alpha|\bm{x}_\alpha)\sim\mathbf{vMF}(\bm{\tilde{\mu}}_{\bm{x}_\alpha},\bm{\tilde{\kappa}}_{\bm{x}_\alpha}),q_\theta(\bm{z}_\beta|\bm{x}_\beta)\sim\mathbf{vMF}(\bm{\tilde{\mu}}_{\bm{x}_\beta},\bm{\tilde{\kappa}}_{\bm{x}_\beta})$,
we measure their distributional overlaps with all pre-set proxies as $\mathcal{O}_{(\alpha,\beta)}^{\bm{\hat{\Upsilon}}}=\{C_d(\bm{\tilde{\kappa}}_{\bm{x}})\exp(\bm{\tilde{\kappa}}_{\bm{x}}\bm{\tilde{\mu}}_{\bm{x}}^T\bm{\hat{\upsilon}}_i),\forall\bm{\hat{\upsilon}}_i\in\bm{\hat{\Upsilon}},\bm{x}\in(\bm{x}_\alpha,\bm{x}_\beta)\}$, which indicate the statistical distances between instance embeddings and proxies.
Since the proxies have been priorly positioned covering the entire manifold with uniformity guarantees, the derived $\mathcal{O}_{\alpha}^{\bm{\hat{\Upsilon}}}$ and $\mathcal{O}_{\beta}^{\bm{\hat{\Upsilon}}}$ can comprehensively characterize the geometric location of instance embeddings.
As such, we suggest $\bm{x}_\alpha$ is semantically correlated with $\bm{x}_\beta$ and their distributional discrepancy should be minor if $\mathcal{O}_\alpha^{\bm{\hat{\Upsilon}}}\doteq\mathcal{O}_\beta^{\bm{\hat{\Upsilon}}}$.
Formally, the condition holds true if $\mathtt{arg\,max_{topk}}(\mathcal{O}_\alpha^{\bm{\hat{\Upsilon}}})=\mathtt{arg\,max_{topk}}(\mathcal{O}_\beta^{\bm{\hat{\Upsilon}}})$.
Let $\mathcal{G}$ denote the collection of instance pairs satisfying the above condition.
In these cases, we aim at minimizing the discrepancy between the two vMF distributions to escalate their semantic affinities.
We resort to modeling the distributional discrepancy with KL divergence and accordingly derive the following optimization objective:
\begin{equation}\label{equ:dis2dis}
	\mathop{\mathtt{min}}\limits_q
	\mathbb{E}_{\mathcal{G}} 
	[-\int_{\mathbb{S}^{d-1}}
	q(\bm{z}|\bm{\tilde{\mu}}_{\bm{x}_\alpha},\bm{\tilde{\kappa}}_{\bm{x}_\alpha})\log(\frac{q(\bm{z}|\bm{\tilde{\mu}}_{\bm{x}_\beta},\bm{\tilde{\kappa}}_{\bm{x}_\beta})}{q(\bm{z}|\bm{\tilde{\mu}}_{\bm{x}_\alpha},\bm{\tilde{\kappa}}_{\bm{x}_\alpha})})d\bm{z}].
\end{equation}
Combined with Eq.\;(\ref{equ:vmf1}) and (\ref{equ:vmf2}), we can derive the corresponding regularization term as follows:
\begin{equation}\label{equ:dis2dis_loss}\begin{aligned}
		\mathcal{L}_{str}(\bm{\tilde{\mu}}_{\bm{x}_\alpha}&,\bm{\tilde{\kappa}}_{\bm{x}_\alpha},\bm{\tilde{\mu}}_{\bm{x}_\beta},\bm{\tilde{\kappa}}_{\bm{x}_\beta})=
		-[\frac{\mathcal{I}_{d/2}(\bm{\tilde{\kappa}}_{\bm{x}_\alpha})}{\mathcal{I}_{d/2-1}(\bm{\tilde{\kappa}}_{\bm{x}_\alpha})}\cdot \\
		(\bm{\tilde{\kappa}}_{\bm{x}_\alpha}-&\bm{\tilde{\kappa}}_{\bm{x}_\beta}\bm{\tilde{\mu}}_{\bm{x}_\alpha}\bm{\tilde{\mu}}_{\bm{x}_\beta}^T)
		+\log(\frac{C_d(\bm{\tilde{\kappa}}_{\bm{x}_\alpha})}{C_d(\bm{\tilde{\kappa}}_{\bm{x}_\beta})})+1]^{-1}.
\end{aligned}\end{equation}
The constraints advocate explicit associations between the geometric layout of embedding manifold and the semantic representations of instances, which in turn systematically structure the open space.
By integrating the optimization objective along with the ones in Eq.\;(\ref{equ:kl_loss}) and (\ref{equ:disperse}), we jointly incorporate the \emph{boundness} and \emph{uniformity} inductive biases to shape the learned embedding space and thereby favor its generalization to unseen biomedical concepts.
The overall training procedure is summarized in the supplementary.

\subsection{Spectral-Based Class Number Estimation}
With the well-structured embedding space, the subsequent target is to estimate the number of classes in unlabeled data and accordingly perform clustering to assign class label for each instance in $\mathcal{D_U}$.
Different from previous methods counting on the results of empirical evaluations and introducing costly recursive clustering \cite{vaze2022generalized}, we propose a spectral graph-theoretic approach based upon the finding that the eigenvalues of graph Laplacians inherently characterize the underlying connectivity of the latent manifold \cite{fiedler1973algebraic}.
Specifically, we construct a weighted graph over all instances of $\mathcal{D_U}$, in which each node corresponds to the embedding $\bm{z}_i$ of a data point, and the edge weights $\mathbf{W}$ model the correlations between pairs of instances as $\{\bm{z}_i\bm{z}_j^T\}$.
Let $\mathbf{\Lambda}$ denote the diagonal matrix with the row-wise sum of $\mathbf{W}$ as its entries,
the optimization objective for graph partitioning with minimized energy is:
\begin{equation}\label{equ:graph_min}
	\mathop{\mathtt{min}}\limits_{\bm{h}}\, \frac{\bm{h}^T\mathbf{\Lambda}^{-1/2}(\mathbf{\Lambda}-\mathbf{W})\mathbf{\Lambda}^{-1/2}\bm{h}}{\bm{h}^T\bm{h}},
\end{equation}
which is equivalent to solve $\mathbf{\Lambda}^{-1/2}(\mathbf{\Lambda}-\mathbf{W})\mathbf{\Lambda}^{-1/2}\bm{h}=\lambda\bm{h}$. The results can be intrinsically depicted by the eigenvectors of the normalized graph Laplacian $\mathbf{\Lambda}^{-1/2}(\mathbf{\Lambda}-\mathbf{W})\mathbf{\Lambda}^{-1/2}$ \cite{boykov2001fast}.
In this regard, the eigenvectors can behave as soft segments to partition the overall graph into a set of semantically-meaningful splits,
while the corresponding eigenvalues can be considered as indicators for the inner-connectivity of potential clusters.
In other words, instances in splits with higher eigenvalues exhibit stronger internal semantic affinities and are more plausible to be assigned into the same class.
We therefore propose to estimate the number of latent classes according to the spectral property of constructed graph that we rank all possible graph splits with their associated eigenvalues and identify the split index which corresponds to the largest gap within the ordering of eigenvalue differences $\{argsort\;\mathtt{diff}(eigenvalues)\}$.
The identified index suggests the start of graph splits which contain limited inner-connectivity and should not be considered as individual classes.
The index is thereby modeled as the estimated number of classes.
Illustrative examples are presented in the supplementary material.
We then perform k-means clustering over $\mathcal{D_U}$ in the hyperspherical space with the estimated class number for label assignment.
Notably, taxonomy-adaptive estimation could be in demand that it is required to gauge the number of potential fine-grained subclasses.
Our method is well suited for this scenario as it can be adapted to estimate the inherent subclasses by loosening the above condition and instead indexing the subordinately largest eigenvalue gap according to the specific need in class hierarchy.

\subsection{Theoretical Analysis}
In this section, we motivate our method theoretically and provide analytical insights to validate its merits.

\noindent\textbf{Monotonicity between Distributional Concentration and Semantic Ambiguity.} \ 
With the intrinsic semantic ambiguity present in biomedical visual data due to disparate imaging protocols across cohorts, our probabilistic treatment, which advocates for vMF statistical modeling, adeptly characterizes the ambiguous attributes with distributional concentration $\tilde{\kappa}$:
\begin{proposition}
	Let $\zeta_{\bm{x}}$ be the continuous entropy of the posterior vMF distribution parametrized by $\bm{\tilde{\mu}_x}\in\mathbb{S}^{d-1}$ and $\bm{\tilde{\kappa}_x}\in\mathbb{R}^{d}_{>0}$.
	We have $\zeta_{\bm{x}}(\bm{\tilde{\kappa}_x})$ behave as a monotonically decreasing function in the interval $(0,+\infty)$.
\end{proposition}
\noindent 
It suggests the statistical parameter $\bm{\tilde{\kappa}_x}$ essentially measures the aleatoric uncertainty of the semantic context in data point $\bm{x}$.
Combined with the analysis of $\bm{\tilde{\kappa}_x}$ on distributional regularization terms, as presented in the supplementary material, it is demonstrated the probabilistic modeling dynamically calibrates the biased learning procedure towards discriminative contextual information modeling.

\noindent\textbf{Towards Bounded Open Space Risk with Uniform Proxies.} \ 
The most challenging aspect in the GNCD problem is how to demystify the dark information regarding unknown class representations in order to restrain the associated open space risk \cite{pu2023dynamic}.
We hereby prove the effectiveness of our proposed method towards a tighter error bound for open space risk characterization and regularization.
Details can be found in the supplemental material.

\newcolumntype{a}{>{\columncolor{light-gray}}c}
\begin{table*}[!t]
	\centering
	\fontsize{8}{9.5}\selectfont
	\begin{threeparttable}
		\caption{Comparison results of our proposed method against other state-of-the-art methods for discovering novel types of pneumonia infectious organisms and cell nuclei. 
			The subscripts denote the standard deviations. The best and second-best results are highlighted in bold and brown, respectively. 
		}
		\label{tab:mainresult1}
		\setlength{\tabcolsep}{0.8mm}{
			\begin{tabular}{c||cca|cca||cca|cca}
				\toprule[0.5mm]
				\multirow{3}{*}{\footnotesize \textbf{Methods}}
				&\multicolumn{6}{c}{\textbf{Pneumonia Infectious Organisms}}
				&\multicolumn{6}{c}{\textbf{Cell Nuclei}}\cr
				\cmidrule(lr){2-7} \cmidrule(lr){8-13} 
				&$\mathsf{Acc}$-all&$\mathsf{Acc}$-known&$\mathsf{Acc}$-novel&$\mathsf{F_1}$-all&$\mathsf{F_1}$-known&$\mathsf{F_1}$-novel&$\mathsf{Acc}$-all&$\mathsf{Acc}$-known&$\mathsf{Acc}$-novel&$\mathsf{F_1}$-all&$\mathsf{F_1}$-known&$\mathsf{F_1}$-novel\cr
				\midrule[0.3mm]				
				RankStats+
				&41.32$_{0.90}$&40.57$_{0.60}$&46.67$_{2.37}$
				&28.96$_{1.56}$&33.45$_{1.13}$&24.48$_{3.13}$
				&36.25$_{0.80}$&36.17$_{0.82}$&39.84$_{0.65}$
				&27.17$_{0.68}$&42.45$_{0.74}$&4.26$_{0.41}$\cr
				UNO+
				&49.59$_{1.48}$&50.06$_{0.92}$&46.31$_{1.49}$
				&31.24$_{1.26}$&36.35$_{2.11}$&26.14$_{2.70}$
				&41.83$_{0.92}$&41.59$_{0.74}$&47.83$_{1.22}$
				&33.04$_{0.58}$&\textbf{45.79}$_{0.69}$&13.54$_{0.52}$\cr
				GCD
				&67.77$_{0.83}$&69.81$_{1.45}$&53.34$_{0.67}$
				&38.06$_{1.04}$&41.26$_{1.30}$&32.87$_{1.95}$
				&45.43$_{1.78}$&45.57$_{1.26}$&43.73$_{2.43}$
				&38.19$_{1.47}$&\textcolor{brown}{44.52$_{1.05}$}&29.64$_{2.84}$\cr
				DCCL
				&\textcolor{brown}{72.73$_{2.14}$}&\textcolor{brown}{74.48$_{2.28}$}&\textcolor{brown}{60.32$_{2.82}$}
				&\textcolor{brown}{44.82$_{1.63}$}&\textbf{49.85}$_{1.95}$&\textcolor{brown}{39.79$_{4.08}$}
				&43.93$_{1.34}$&43.59$_{1.86}$&54.49$_{0.53}$
				&37.93$_{1.14}$&40.16$_{2.04}$&\textcolor{brown}{36.57$_{0.77}$}\cr
				\midrule[0.2mm]						
				SLF
				&43.81$_{0.67}$&42.31$_{0.44}$&54.26$_{1.12}$
				&30.72$_{1.39}$&32.86$_{0.97}$&28.57$_{2.86}$
				&39.15$_{0.86}$&39.07$_{1.04}$&43.06$_{0.69}$
				&28.95$_{0.95}$&44.06$_{1.02}$&18.42$_{0.81}$\cr
				BYOP
				&65.52$_{0.95}$&68.87$_{1.14}$&58.95$_{2.54}$
				&40.01$_{1.02}$&43.12$_{1.30}$&38.52$_{3.72}$
				&\textcolor{brown}{46.28$_{0.68}$}&\textcolor{brown}{46.02$_{0.57}$}&\textcolor{brown}{55.82$_{1.44}$}
				&\textcolor{brown}{39.06$_{0.59}$}&41.01$_{0.30}$&35.63$_{1.86}$\cr
				\midrule[0.2mm]	
				\textbf{Ours}
				&\textbf{78.51}$_{1.05}$&\textbf{80.19}$_{1.02}$&\textbf{66.67}$_{2.70}$
				&\textbf{46.73}$_{1.47}$&\textcolor{brown}{44.27$_{0.31}$}&\textbf{49.18}$_{3.87}$
				&\textbf{47.78}$_{1.28}$&\textbf{47.41}$_{1.53}$&\textbf{57.85}$_{0.42}$
				&\textbf{42.64}$_{0.36}$&41.48$_{0.25}$&\textbf{43.27}$_{2.29}$\cr	
				\bottomrule[0.5mm]
		\end{tabular}}
	\end{threeparttable}
\end{table*}
\begin{table*}[!t]
	\centering
	\fontsize{8}{9.5}\selectfont
	\begin{threeparttable}
		\caption{Comparison results for discovering novel types of skin cancer lesions and degrees of diabetic retinopathy severity. 
		}
		\label{tab:mainresult2}
		\setlength{\tabcolsep}{0.8mm}{
			\begin{tabular}{c||cca|cca||cca|cca}
				\toprule[0.5mm]
				\multirow{3}{*}{\footnotesize \textbf{Methods}}
				&\multicolumn{6}{c}{\textbf{Skin Cancer Lesions}}
				&\multicolumn{6}{c}{\textbf{Diabetic Retinopathy Severity}}\cr
				\cmidrule(lr){2-7} \cmidrule(lr){8-13} 
				&$\mathsf{Acc}$-all&$\mathsf{Acc}$-known&$\mathsf{Acc}$-novel&$\mathsf{F_1}$-all&$\mathsf{F_1}$-known&$\mathsf{F_1}$-novel&$\mathsf{Acc}$-all&$\mathsf{Acc}$-known&$\mathsf{Acc}$-novel&$\mathsf{F_1}$-all&$\mathsf{F_1}$-known&$\mathsf{F_1}$-novel\cr
				\midrule[0.3mm]				
				RankStats+
				&29.19$_{1.03}$&28.65$_{0.79}$&33.39$_{2.04}$
				&21.06$_{1.60}$&36.43$_{0.95}$&9.53$_{2.84}$
				&47.21$_{0.94}$&47.90$_{0.58}$&41.64$_{1.50}$
				&40.82$_{0.69}$&54.78$_{0.46}$&19.97$_{0.82}$\cr
				UNO+
				&37.80$_{2.01}$&37.43$_{1.54}$&\textcolor{brown}{40.74$_{2.38}$}
				&29.16$_{1.49}$&40.68$_{1.41}$&20.37$_{1.82}$
				&50.43$_{0.85}$&51.07$_{0.97}$&39.26$_{0.68}$
				&36.57$_{0.84}$&52.45$_{0.57}$&16.09$_{1.23}$\cr
				GCD
				&31.58$_{1.44}$&49.46$_{1.29}$&39.58$_{1.06}$
				&27.18$_{0.82}$&39.17$_{0.53}$&18.20$_{2.25}$
				&48.43$_{2.36}$&49.05$_{2.04}$&44.33$_{1.22}$
				&41.69$_{1.06}$&53.39$_{0.85}$&21.13$_{0.93}$\cr
				DCCL
				&56.45$_{0.78}$&58.92$_{0.43}$&37.84$_{1.37}$
				&33.46$_{1.14}$&53.46$_{0.59}$&18.48$_{1.50}$
				&44.35$_{1.58}$&44.93$_{1.44}$&\textcolor{brown}{46.07$_{2.27}$}
				&43.23$_{1.55}$&52.16$_{1.40}$&\textcolor{brown}{27.51$_{1.89}$}\cr
				\midrule[0.2mm]						
				SLF
				&44.56$_{1.29}$&45.14$_{0.73}$&39.02$_{0.67}$
				&30.78$_{1.77}$&49.75$_{0.84}$&16.55$_{1.96}$
				&46.29$_{0.99}$&46.93$_{0.65}$&44.40$_{1.40}$
				&40.41$_{1.13}$&50.92$_{0.54}$&20.10$_{1.72}$\cr
				BYOP
				&\textcolor{brown}{65.79$_{0.62}$}&\textcolor{brown}{69.46$_{0.44}$}&36.65$_{1.51}$
				&\textcolor{brown}{35.94$_{1.14}$}&\textcolor{brown}{54.19$_{0.71}$}&\textcolor{brown}{23.04$_{1.70}$}
				&\textcolor{brown}{51.67$_{1.24}$}&\textcolor{brown}{52.31$_{1.27}$}&43.63$_{0.93}$
				&\textcolor{brown}{44.85$_{0.82}$}&\textcolor{brown}{55.68$_{1.23}$}&25.56$_{0.90}$\cr
				\midrule[0.2mm]	
				\textbf{Ours}
				&\textbf{69.62}$_{0.96}$&\textbf{73.24}$_{0.32}$&\textbf{41.67}$_{0.55}$
				&\textbf{41.39}$_{0.29}$&\textbf{60.41}$_{0.40}$&\textbf{27.12}$_{0.48}$
				&\textbf{58.41}$_{1.35}$&\textbf{59.62}$_{1.60}$&\textbf{50.52}$_{0.51}$
				&\textbf{47.25}$_{0.77}$&\textbf{55.94}$_{0.56}$&\textbf{34.22}$_{0.64}$\cr	
				\bottomrule[0.5mm]
		\end{tabular}}
	\end{threeparttable}
\end{table*}
\section{Experiments}
\subsection{Experimental Setup}
\noindent\textbf{Datasets.} \ 
To evaluate the effectiveness and general applicability of our method under miscellaneous biomedical discovery scenarios, we conduct experiments on a diverse suite of challenging tasks encompassing various biomedical concepts\;(cell, lesion, disease), imaging modalities\;(X-ray, microscopy, dermatoscopy, fundus photography), and pathologies\;(lung infection, skin cancer, eye complication).
For each task, we adopt two different datasets with evident data distribution shifts to serve as the labeled base set and the unlabeled set comprising novel classes, respectively.
In specific, we consider the discovery of novel types or degrees of pneumonia infectious organisms \cite{kermany2018identifying, cohen2020covid}, cell nuclei \cite{gamper2019pannuke, graham2021lizard}, skin lesions \cite{tschandl2018ham10000, codella2019skin}, and diabetic retinopathy severity \cite{li2019diagnostic, aptos2019}.
The quantity of instances belonging to novel classes is inherently much smaller compared with their base counterpart and naturally formalize the long-tailed class distribution.
More details can be found in the supplemental material,
where extended studies on general concept discovery are further presented and analyzed.

\noindent\textbf{Implementation Details and Evaluation Protocol.} \ 
We adopt the ViT-B/16 \cite{dosovitskiy2021an} with DINO pre-trained weights \cite{caron2021emerging} as the backbone network for fair comparisons with the previous methods \cite{vaze2022generalized, pu2023dynamic}.
Its output of $\mathtt{[CLS]}$ token with a dimension of 768 is used for latent embedding.
Please refer to the attached code for details.

For evaluation, we measure the clustering accuracy between the predicted class assignment and the ground-truth label.
Following the protocol in \cite{vaze2022generalized}, we match the two class sets by searching for the optimal permutation using the Hungarian assignment algorithm \cite{kuhn1955hungarian}.
We adopt the accuracy and F1 scores as metrics and report the averaged score across the entire class set.
Particularly, we also present the scores specific to the base known $\mathcal{S_B}$ and novel $\mathcal{S}_N$ subset of classes in a separated manner.

\begin{figure}[!t]
	\centerline{\includegraphics[width=\columnwidth]{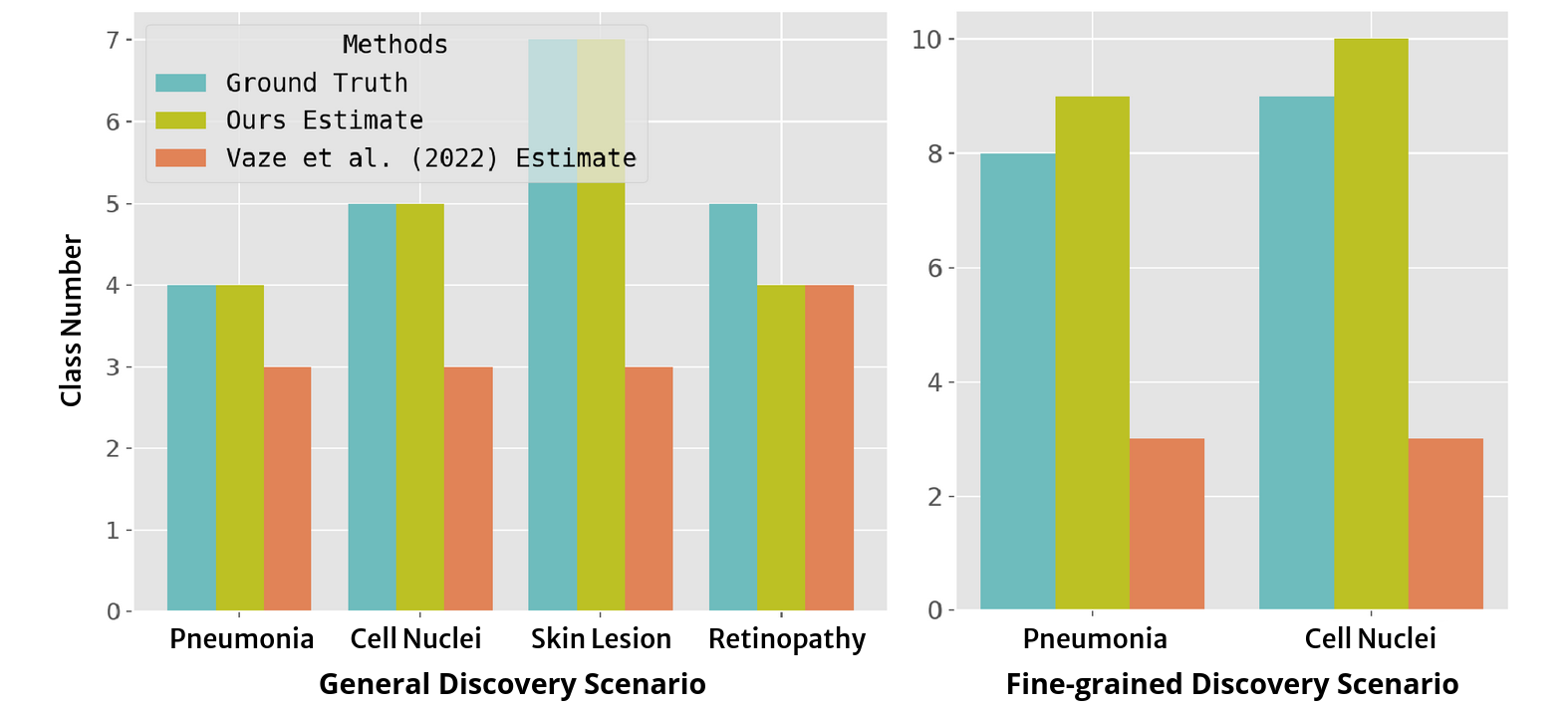}}
	\caption{Results of class number estimation in the unlabeled set. For pneumonia and cell nuclei, we further estimate the number of fine-grained subclasses.}
	\label{fig:estimate}
\end{figure}
\subsection{Results and Discussions}
\noindent\textbf{Main Results.} \ 
We compare our method with state-of-the-art approaches for novel class discovery\;(NCD), including RankStats+ \cite{han2021autonovel}, UNO+ \cite{fini2021unified}, GCD \cite{vaze2022generalized}, and DCCL \cite{pu2023dynamic} which are specifically devised for the challenging generalized NCD setting, as well as SLF \cite{yu2022self} and BYOP \cite{yang2023bootstrap} which take account of the non-i.i.d. data and non-uniform label distribution issues, respectively.
The overall quantitative results over four diverse biomedical concept discovery tasks are presented in Table\;\ref{tab:mainresult1} and \ref{tab:mainresult2}.
All methods are implemented based upon the same ViT-B/16 backbone with DINO pre-trained weights \cite{vaze2022generalized}.
It can be observed that our proposed method consistently and significantly outperforms the state-of-the-art novel class discovery approaches over all four tasks.
Notably, as highlighted in the gray columns which correspond to the clustering metrics specific to novel classes, our method surpasses the comparison approaches by a substantial margin especially on the identification and grouping of unseen phenotypes.
The results demonstrate the advantages and effectiveness of the proposed geometry-constrained probabilistic modeling for discovering novel concepts despite the data and label distribution biases intrinsically present in the biomedical domain.

We further assess the efficacy of the spectral-based class number estimation method, as presented in Fig.\;\ref{fig:estimate}.
The proposed approach attains promising results that the estimated number is identical or very close to the ground-truth value.
The versatile method also holds an appealing merit that it supports taxonomy-adaptive estimation, which means it is capable of estimating the number of potential fine-grained subclasses, a practical aspect not previously considered in the literature at all.
Additionally, it is noteworthy that our method does not rely on recursive clustering as in \cite{vaze2022generalized} and is therefore computationally efficient.
On the pneumonia benchmark, our method only takes \textbf{0.121s}, which is more than $\mathbf{10\times}$ faster than the comparison approach \cite{vaze2022generalized} consuming \textbf{2.854s}.

\begin{table}[!t]
	\centering
	\fontsize{7.6}{8.5}\selectfont
	\begin{threeparttable}
		\caption{Key component analysis of our method.
			$\mathtt{Uni}$, $\mathtt{Bnd}$, $\mathtt{Disp}$, and $\mathtt{Str}$ correspond to uniform proxies, base space bounding, open space dispersion and structuring, respectively.
		}
		\label{tab:ablation}
		\setlength{\tabcolsep}{0.6mm}{
			\begin{tabular}{cccc||cca|cca}
				\toprule[0.5mm]
				\multicolumn{4}{c}{\textbf{Components}}				&\multicolumn{3}{c}{\textbf{Pneumonia}}
				&\multicolumn{3}{c}{\textbf{Cell Nuclei}}\cr
				\cmidrule(lr){1-4} \cmidrule(lr){5-7} \cmidrule(lr){8-10} 
				$\mathtt{Uni}$&$\mathtt{Bnd}$&$\mathtt{Disp}$&$\mathtt{Str}$&$\mathsf{F_1}$-all&$\mathsf{F_1}$-known&$\mathsf{F_1}$-novel&$\mathsf{F_1}$-all&$\mathsf{F_1}$-known&$\mathsf{F_1}$-novel\cr
				\midrule[0.3mm]				
				&&&
				&22.82&27.37&18.28
				&16.21&20.76&8.64\cr				
				$\checkmark$&$\checkmark$&&
				&35.84&\textbf{48.14}&23.53
				&34.30&38.76&27.72\cr			
				$\checkmark$&$\checkmark$&$\checkmark$&
				&36.92&45.89&27.94
				&33.02&31.83&33.47\cr		
				$\checkmark$&$\checkmark$&&$\checkmark$
				&28.71&36.51&20.91
				&38.23&40.49&35.59\cr	
				&$\checkmark$&$\checkmark$&$\checkmark$
				&40.23&37.69&31.42
				&31.97&34.85&29.74\cr
				\midrule[0.2mm]	
				$\checkmark$&$\checkmark$&$\checkmark$&$\checkmark$
				&\textbf{46.73}&44.27&\textbf{49.18}
				&\textbf{42.64}&\textbf{41.48}&\textbf{43.27}\cr	
				\bottomrule[0.5mm]
		\end{tabular}}
	\end{threeparttable}
\end{table}
\noindent\textbf{Ablation of Key Components.} \ 
Table\;\ref{tab:ablation} presents the ablation study results to demonstrate the effectiveness of key components in our proposed method.
It can be remarked that each component is indispensable and holds a positive impact on novel class discovery.
Notably, the combination of pre-defined uniform proxies and base space bounding reaches the peak accuracy in terms of $\mathsf{F_1}$-known on the pneumonia infectious organism benchmark.
However, it suffers from inferior efficacy in identifying and clustering potential novel concepts.
The observation is in line with our argument that the consideration and explicit modeling of the dark information regarding unknown classes are crucial to ensure the generalizability of learned representations for discovering unseen concepts.
Solely imposing constraints on the base space is insufficient to calibrate the learning process biased towards the dominant known classes.
In contrast, our method proposes two supplementary geometric constraints which cooperatively shape the layout of embedding space and compensate for the open space risk, which in turn leads to unbiased and discriminative modeling of novel class representations.

\begin{figure}[!t]
	\centerline{\includegraphics[width=1.1\columnwidth]{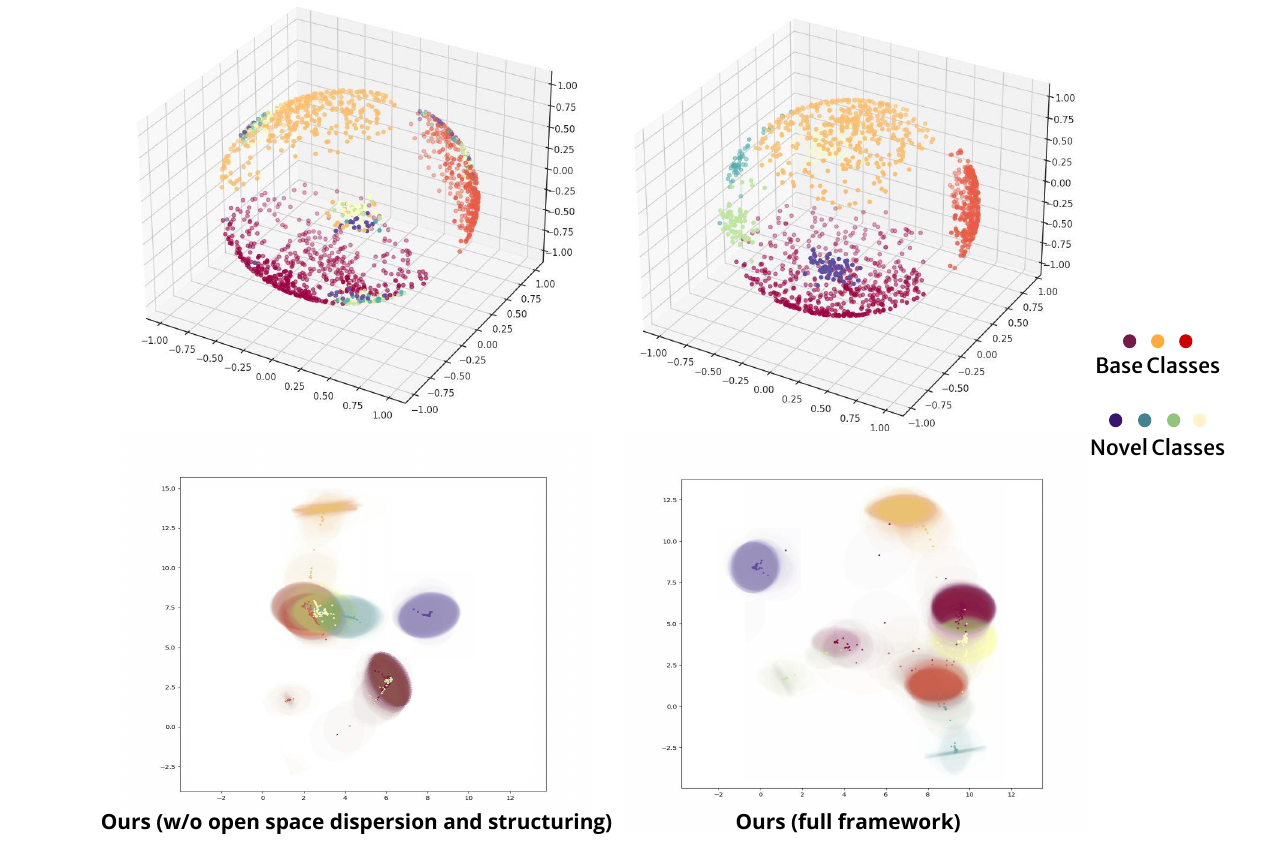}}
	\caption{Hyperspherical and distributional embeddings of the skin lesion test set for two variants of our proposed method. For distributional visualization, we overlay data points over the super-level-set ellipses of the associated probabilistic distributions.}
	\label{fig:embed_vis}
\end{figure}
\noindent\textbf{Impacts on Geometric Layout of Embedding Space.} \ 
To perspicuously illustrate the effectiveness of our method towards shaping the geometric structure of embedding space, we visualize the learned representations on the 3D spherical manifold and its associated probabilistic characterizations, as shown in Fig.\;\ref{fig:embed_vis}.
The covariance structure of data points is modeled as a group of super-level-set ellipses to indicate the uncertainty of constructed clusters.
We observe that the removal of open space dispersion and structuring inevitably deteriorates the degree of inter-class angular separation and results in ambiguous semantic representations.
The comparative analysis between the two variants of our proposed method substantiates the significance of incorporating geometric inductive biases to regulate the underlying organization of the embedding space.

\section{Conclusion}
In this work, we propose a geometry-constrained probabilistic modeling framework to address challenging aspects of novel class discovery in the biomedical domain.
Specifically, we suggest to model the latent representation of each instance as a directional distribution to exclude the task-irrelevant data variations arising from inconsistent imaging protocols across cohorts.
We then identify several geometric properties critical for organic embedding space structuring. 
By explicitly integrating those principles to regularize representation layout, the open space risk associated with the discovery of novel classes can be effectively bounded.
Additionally, a spectral graph-theoretic algorithm is developed to estimate the number of categories in unlabeled data, which inherits taxonomy adaptivity and high efficiency.
Our method exhibits appealing performance on a diverse suite of novel concept discovery applications in biomedical domains.

{\small
\bibliographystyle{ieee_fullname}
\bibliography{egbib}}

\end{document}